\documentclass[10pt,twocolumn,letterpaper]{article}

\usepackage{iccv}              %

\definecolor{iccvblue}{rgb}{0.21,0.49,0.74}
\usepackage[pagebackref,breaklinks,colorlinks,allcolors=iccvblue]{hyperref}

\usepackage{orcidlink}
\usepackage{adjustbox}
\usepackage{multirow}
\usepackage{multicol}
\usepackage{arydshln}
\usepackage{amsmath}
\usepackage{xcolor}
\usepackage{fontawesome}
\usepackage{graphicx}

\newcommand{\boldparagraph}[1]{\vspace{0.3em}\noindent{\bf #1} }

\newcommand{\alignment}{alignment-cluster }

\title{Multi Activity Sequence Alignment via Implicit Clustering}

\author{Taein Kwon$^1$~~~~~~~~~~~~~~~~~~~
Zador Pataki$^1$~~~~~~~~~~~~~~~~~~~
Mahdi Rad$^2$~~~~~~~~~~~~~~~~~~~
Marc Pollefeys$^{1,2}$
\smallskip 
\\
$^1$ETH Z\"urich~~~~~~$^2$Microsoft MR \& AI Lab, Z\"urich
}

\begin{document}

\twocolumn[{%
  \renewcommand\twocolumn[1][]{#1}%
  \maketitle
  \vspace{-5mm}
  \includegraphics[width=\textwidth]{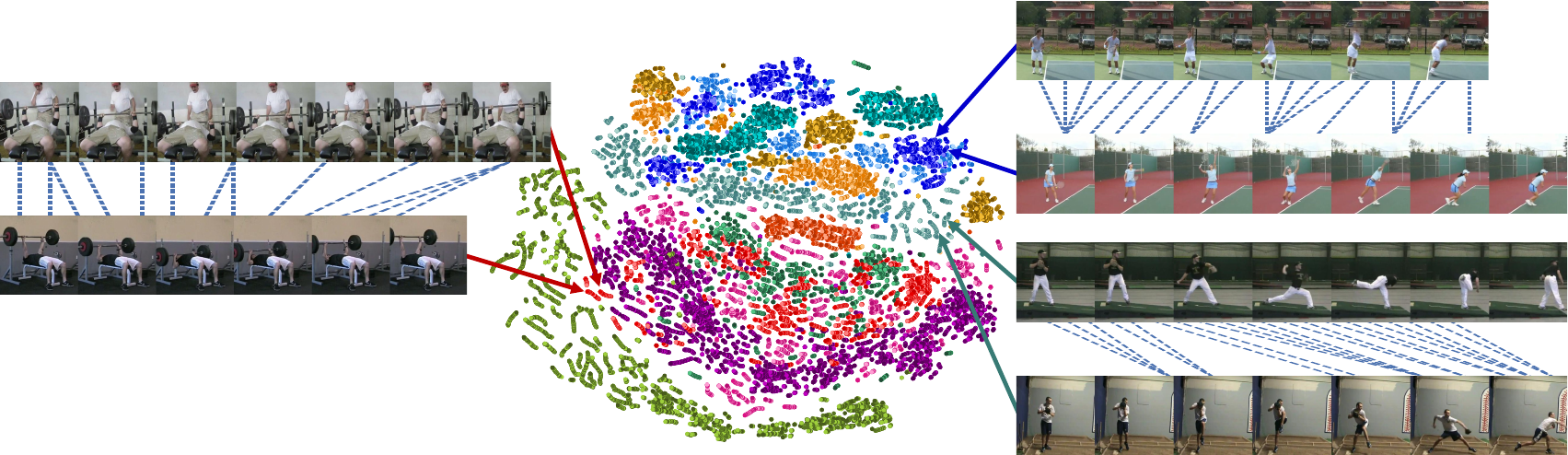}
  \captionof{figure}{\textbf{The t-SNE visualization of the learned embeddings on the PennAction dataset.} The proposed method effectively distinguishes different activities. On the left, the alignment of two bench pressing sequences is shown, which demonstrates the effectiveness of our approach. On the right, our model not only differentiates between activities like baseball pitch and tennis serve but also recognizes their proximity in the embedding domain. Dashed lines indicate matched frames between sequences.}
  \label{fig:teaser}
  
  \vspace{5mm}
}]
\begin{abstract}

Self-supervised temporal sequence alignment can provide rich and effective representations for a wide range of applications. 
However, existing methods for achieving optimal performance are mostly limited to aligning sequences of the same activity only and require separate models to be trained for each activity. 
We propose a novel framework that overcomes these limitations using sequence alignment via implicit clustering. 
Specifically, our key idea is to perform implicit clip-level clustering while aligning frames in sequences. This coupled with our proposed dual augmentation technique enhances the network's ability to learn generalizable and discriminative representations.
Our experiments show that our proposed method outperforms state-of-the-art results and highlight the generalization capability of our framework with multi activity and different modalities on three diverse datasets, H2O, PennAction, and IKEA ASM. We will release our code upon acceptance.
\end{abstract}
    
\section{Introduction}
\label{sec:intro}

Dense temporal alignment of videos~\cite{dwibedi2019temporal,sermanet2018time} is a fundamental task in computer vision, which aims to find the optimal frame-wise correspondence between two or more video sequences. It has many applications, such as frame retrieval, action recognition, abnormality detection, and skill transfer. Moreover, the advancements in AR and VR have created new opportunities for utilizing temporal alignment in understanding human activity, supported by diverse modalities such as RGB images, hand movements, and body poses.

To learn dense temporal alignment, supervised learning-based methods require large amounts of labeled data for each activity category and alignment. However, manually labeling data is expensive, limiting the generalizability and scalability of these methods to new activities and domains. Hence, there has been a focus on using self-supervised learning methods, such as cycle consistency~\cite {dwibedi2019temporal} and temporal alignment~\cite{haresh2021learning}, to perform dense video alignment.
However, when extending these methods to encompass \emph{multiple activities}, they require action labels as their nature are primarily designed for videos of the same activity. Consequently, they either operate solely on single-activity videos or require a separate model to be trained for each activity.

Unlike previous methods, contrastive learning-based methods, on RGB~\cite{chen2022frame, sermanet2018time} or on appearance-invariant data such as skeletal~\cite{kwon2022context, tran2023learning} can be trained on multiple activities without requiring any additional labels.
However, as our experiments show, when these methods are trained on multiple activities, the learned embeddings tend to collapse into a single homogeneous space across different activities, resulting in difficulties in distinguishing between different activities, and therefore limiting the applicability of temporal video alignment in real-world scenarios.

In this paper, we introduce a novel framework for self-supervised sequence alignment that overcomes these limitations by leveraging the power of implicit clustering techniques. We name it MASA (Multi Activity Sequence Alignment). As illustrated in Figure~\ref{fig:teaser}, MASA is not only able to learn discriminative representations between different activities but also recognize the proximity of similar activities in the embedding domain. MASA consists of three main components: the augmentation module, the context-aware module, and the \alignment module. 
The augmentation module applies our augmentation techniques to each sequence to perform contrastive style learning. The context-aware module then extracts embeddings from the input sequences, capturing rich temporal information. Finally, the novel \alignment module performs implicit clustering and matching of frames between input sequences. 
Inspired by \cite{chen2021exploring}, originally designed for image classification tasks, 
we extend the concept to clip-level clustering by leveraging the attention mechanism, to map each sequence to a latent vector that represents its clustering prototype via the clip-level clustering predictor.
In addition to the novel \alignment module, we propose a dual augmentation technique that aids in learning more generalizable and discriminative representations than the existing methods~\cite{kwon2022context,tran2023learning}. The existing techniques lack the diversity of temporal variations, whereas ours provides a wider range. By jointly optimizing these modules, MASA can cluster and align sequences in a self-supervised fashion.
  
We conduct a thorough evaluation of the effectiveness of our approach by assessing its performance in various downstream tasks related to fine-grained video understanding, such as phase classification, phase progress, Kendall tau, and frame retrieval, on three diverse datasets: H2O \cite{Kwon_2021_H2O}, PennAction \cite{Zhang_2013_Penn}, and IKEA ASM \cite{Ben-shabat_2020_Ikea}, which contain diverse temporal range from a few seconds to minutes for activities. 
As light representations of human/hand poses and motions such as 3D skeletons are beneficial for AR and VR scenarios, we show our proposed method can perform well on these representations and it is not limited to RGB images only. Moreover, 3D skeletons can be obtained easily thanks to recent off-the-shelf methods and often from AR and VR devices.
Furthermore, we show the clustering ability of the proposed network by performing action recognition across different activities. 
Our method can leverage multiple modalities, and results in even better accuracy.

Our main contributions are summarized as follows:
\begin{itemize}
\item	We introduce MASA, a novel framework for multi activity sequence alignment, which can effectively handle a wide range of activities and scenarios. %

\item We propose implicit clip-level clustering, which identifies prototypes of different activities in the latent space. To the best of our knowledge, we are the first to leverage implicit clustering while optimizing sequence alignment.
With our proposed dual augmentation technique, the implicit clip-level clustering enables the framework to learn more generalizable and discriminative representations.

\item We demonstrate MASA is not limited to RGB only, and it can perform on different modalities such as skeleton, showing its generalization ability.
\item We prove MASA achieves state-of-the-art on most of the metrics on both modalities, RGB and skeleton. 
Notably, MASA is not only able to handle multiple activities to train on a single model,
 but it also leverages the diverse activity data to boost overall performance even further. Previous approaches, however, suffer performance degradation with multiple activities on a single model.

\end{itemize}
We demonstrate the superiority of MASA on three diverse datasets and provide ablation studies and quantitative analysis as well as showing clustering performance through the action recognition task for both RGB and skeleton inputs.

\section{Related Work}
\label{sec:related_work}
\boldparagraph{Self-Supervised Representation Learning.} 
Driven by the need to scale up training by leveraging unlabeled data, self-supervised learning (SSL) has become a crucial method in vision analysis. Pioneering works explored contrastive learning approaches to learn image representations in a self-supervised manner. SimCLR~\cite{chen2020simple} demonstrated the benefits of data augmentation and contrasting representations in a remapped latent space, while \cite{wu2018unsupervised} and MoCo~\cite{he2020momentum, chen2020improved} identified performance improvements resulting from a large number of negative anchors. 
Recent works explored the benefits of combining clustering and representation learning~\cite{asano2019self, caron2020unsupervised, chen2021exploring}, and BYOL~\cite{grill2020bootstrap} and SimSiam~\cite{chen2021exploring} introduced implicit clustering with contrastive learning approaches that do not rely on negative samples. Besides contrastive learning, MAE~\cite{he2022masked} and DINO~\cite{caron2021emerging} learn effective image representations using masked autoencoding and self-distillation respectively. 
Contrastive learning approaches have also been extended to learn video-level representations~\cite{dave2022tclr, feichtenhofer2021large, hu2021contrast, qian2021spatiotemporal}. Inspired by the field of Natural Language Processing, future frame prediction has been explored~\cite{ahsan2018discrimnet, diba2019dynamonet} and MAE inspired a series of works in video-level representation learning~\cite{tong2022videomae, wang2023videomae}. In this work, we explore the pretext task of sequence alignment as an SSL approach for clip-level representation learning. Furthermore, we demonstrate that employing clip-level implicit clustering can generalize to multi activity sequence alignment.

\boldparagraph{Sequence Alignment.} 
Sequence alignment in video analysis refers to the process of aligning sequences of frames or actions across different videos. Early methods, such as TCN~\cite{sermanet2018time}, utilized contrastive learning on synchronized frames across views. This method, however, lacks scalability due to the need for a multi-camera setup. TCC~\cite{dwibedi2019temporal} introduced scalable SSL through temporal cycle-consistency for robust frame correspondence. Subsequent developments like LAV~\cite{haresh2021learning} and VAVA~\cite{liu2022learning} adopted dynamic time warping and optimal transport for SSL in video alignment. These methods, however, require action labels when they train all activities in one model. 
Alternatively, CARL~\cite{chen2022frame} extracts random clips from a video, while VSP~\cite{zhang2023modeling} employs stochastic process modeling to enable contrastive learning. 
For skeleton features, CASA~\cite{kwon2022context}, using sequence augmentation and transformer-based attention, significantly improved frame-wise embeddings for 3D skeletons, with LA2DS~\cite{tran2023learning} following suit. 
However, these augmentation-based methods~\cite{kwon2022context,tran2023learning,chen2022frame}, stochastic modeling method~\cite{zhang2023video} 
are not dedicatedly designed for multi activity scenarios and often face limitations in action-specific generalization. More recently, GTCC~\cite{donahue2024learning} employs the Gaussian mixture model and context-dependent drop function to generalize TCC~\cite{dwibedi2019temporal}. However, their performance is suboptimal compared to other methods~\cite{liu2022learning,chen2022frame}.
MASA overcomes this by employing implicit clustering to train networks capable of generating action-generalizable embeddings as well as distinguishing across different activities. %

\boldparagraph{Action Recognition.} 
Action recognition is a task in computer vision that involves identifying and classifying human actions based on a sequence of data. While supervised methods~\cite{carreira2017quo,feichtenhofer2019slowfast} were widely used before, with the development of large-scale video datasets, video representation learning~\cite{tong2022videomae,wang2023videomae,oquab2023dinov2,qian2021spatiotemporal, Han20}, which treats action recognition as downstream tasks, has gained popularity. 
In AR and VR scenarios, skeletal action recognition has drawn significant attention~\cite{yan2018spatial, shi2019two, cheng2020skeleton, liu2020disentangling, lin2023casar,wen2023hierarchical,cho2023transformer} due to its light representation, especially as mixed reality devices often provide skeletons prediction.
Similar to video representation learning with RGB data, recent advancements in skeletal action recognition have explored SSL methods, including contrastive learning \cite{su2021self} and using future pose prediction as a pretext task \cite{li2019actional}. Furthermore, clustering has helped improve representation qualities \cite{su2020predict, kumar2022unsupervised}, and node-level MAE has been deployed in skeleton sequences \cite{yan2023skeletonmae, mao2023masked}. 
We demonstrate the effectiveness of our SSL sequence representation on multi activity by implicit clustering in action recognition benchmarks.

\begin{figure*}
   \includegraphics[width=1.00\textwidth]{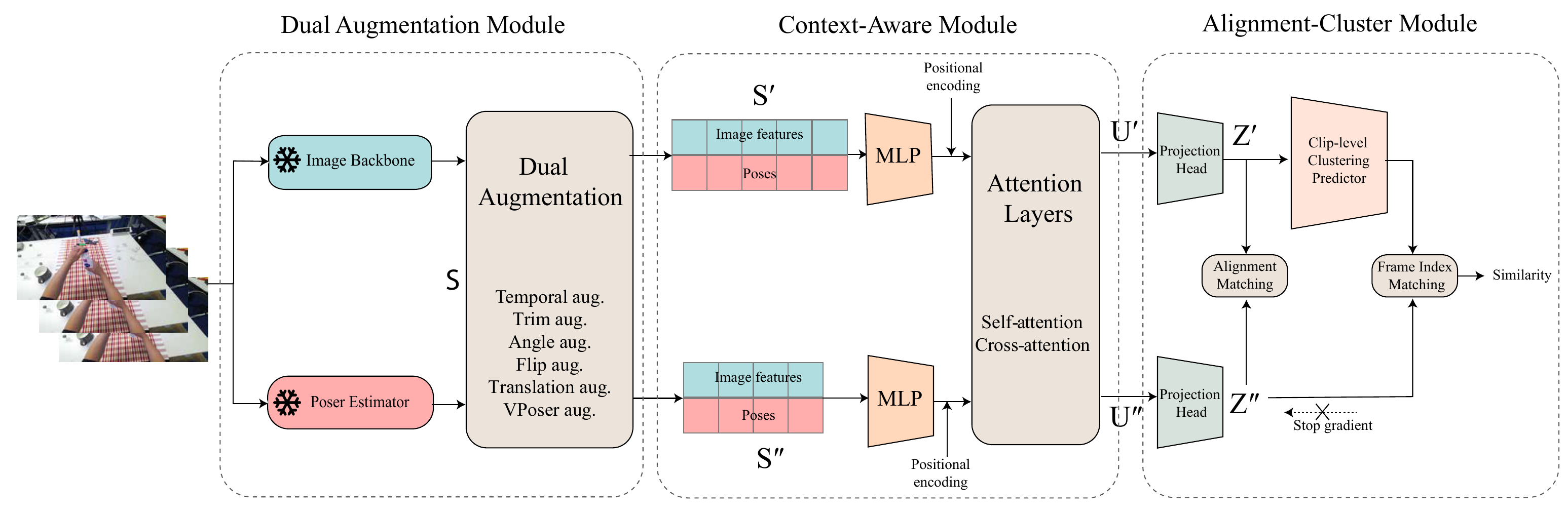}
   \vspace{-8mm}
   \caption{\textbf{System Overview}. Our framework takes RGB images as inputs and obtains RGB features using the frozen DINO~\cite{oquab2023dinov2} pre-trained model and skeletons extracted using FrankMocap~\cite{rong2021frankmocap}. Dual augmentation generates two different augmented sequences with frame-wise concatenated image features and skeletons. Note that when we use only one modality, we disable the other modality and feed it to the context-aware module without concatenation. Our context-aware module extracts embeddings for the downstream tasks. The alignment module matches and clusters the latent features to learn framewise video embeddings.}
   \label{fig:system_overview}
   \vspace{-4mm}
\end{figure*}

\section{Method}
We present the details of our proposed framework for self-supervised sequence alignment as shown in Figure~\ref{fig:system_overview}. MASA consists of three main components: the augmentation module, which performs dual augmentation for the given sequence at training time, the context-aware module which extracts embedding from the input sequences, and the alignment-cluster module which implicitly clusters and matches the frames between the input sequences.

\subsection{Dual Augmentation}
\label{sec:dual_augmentation}

Existing methods~\cite{kwon2022context,tran2023learning} use a temporal augmentation technique such that given a sequence, they generate one augmented sequence for alignment by randomly dropping a number of frames. However, this is suboptimal as it limits the models to learn alignment sequences with faster speed only. We, therefore, propose to extend this technique to Dual Augmentation. Our technique aims to generate two different augmented sequences $\mathcal{S'}=\{s'_j\}^M$ and $\mathcal{S''}= \{s''_k\}^N$ from the same original sequence $\mathcal{S}$ with the random trimming length M and N, where $s'_j$ and $s''_k$ are individual frames of the corresponding sequences. We then randomly trim the original sequence and apply individual temporal augmentation for each sequence, $\mathcal{S'}$ and $\mathcal{S''}$, ensuring that the sequence range remains consistent during clustering. In addition to the advantage of supporting both faster and slower sequences, dual augmentation increases the diversity and complexity of the input sequences, which can help the network learn more generalizable and discriminative representations. This, however, requires a novel matching loss function, which we will present in Section~\ref{sec:alignment_module}.
Moreover, the random trimming technique, while augmenting the sequence, helps to learn partially trimmed sequence clustering without knowing the entire sequence information, leading to a better clustering performance together with the clustering predictor.
Furthermore, dual augmentation allows the network to handle cases where some frames from the original sequence are not present in either of the augmented sequences, which can happen due to random dropping. This can enhance the network's ability to recover missing information and fill in the gaps. Therefore, while augmenting, we compute $g'(\cdot)$ and $g''(\cdot)$, which map indices from $\mathcal{S'}$ and $\mathcal{S''}$ to their original index in $\mathcal{S}$, respectively.
Note that we only perform this trim augmentation for RGB images because the 3D skeleton representation is already compact, and additional trimming results in less context information.

In practice, for spatial augmentation, we follow the augmentation techniques proposed in~\cite{tran2023learning} such as angle, flip, translation and VPoser augmentations. For RGB augmentation, we only perform dual augmentation temporally. We provide a detailed explanation of these techniques in the supplementary material.  

\vspace{5mm}
\subsection{Context-Aware Module}
The context-aware module aims to extract the embeddings from the input sequences that capture the temporal information of the frames. The input sequences are obtained by applying two different augmentations to the original video sequence.
The module first extracts a feature representation of each frame using an MLP model and then employs temporal positional encoding to embed temporal context. We employ sine and cosine function-based positional encoding by following the success of the Transformer~\cite{vaswani2017attention}. We then pass this information to the encoder, which consists of self- and cross-attention encoders. Self-attention layers learn about the relation between frames within the same sequence, while cross-attention layers learn the relation between frames across augmented sequences. 
The output of the context-aware module is a set of embeddings that encode both the spatial and temporal information of the input sequences as follows:

\begin{equation}
(\mathcal{U'}, \mathcal{U''})=A(f(\mathcal{S'}) + \text{PE}_\mathcal{S'},f(\mathcal{S''})+ \text{PE}_\mathcal{S''}), 
\end{equation}
where $f(\cdot)$ is the MLP model described above, $\text{PE}$ is the positional encoding, and $A(\cdot)$ is the attention-based encoder.

\subsection{Alignment-Cluster Module}
\label{sec:alignment_module}
The alignment module aims to find the optimal temporal alignment between the input sequences. 
It takes the embeddings from the context-aware module as inputs and matches them framewise to learn alignment between two augmented sequences.
In practice, we first employ Projection head to the embeddings computed by the context-aware module, which is suggested by SimCLR~\cite{chen2020simple}. This head prevents the embedding space from overfitting to the target tasks, such as frame alignment and clustering, and makes the representation perform downstream tasks easier.

\paragraph{Matching.}
The matching algorithm is a key component of the alignment module. It learns to align two augmented sequences by minimizing the distance between their indices with respect to the original sequence, in a regression-based fashion. MASA is designed in bi-direction unlike the previous methods~\cite{kwon2022context,tran2023learning}, which only consider uni-direction to calculate distance, thanks to the dual augmentation.

More specifically, let $\mathcal{Z'}$ and $\mathcal{Z''}$ be the output of the projection head $P(\cdot)$ given the computed embeddings of the sequences  $\mathcal{U'}$ and $\mathcal{U''}$, respectively. Following~\cite{dwibedi2019temporal, kwon2022context,tran2023learning}, we then calculate $\gamma_{jk}$  the probability of a frame index $j$, where frame $s_j' \in \mathcal{S'}$, being a match, to frame index $k$, where $s_k'' \in \mathcal{S''}$, as 

\begin{equation}\label{eq:gamma}
\gamma_{j,k} = \frac{\text{exp}(\mathcal{D}(\mathcal{Z}_j',\mathcal{Z}_k''))}
{\sum_{n=1}^{N}\text{exp}(\mathcal{D}(\mathcal{Z}_j',\mathcal{Z}_n''))},
\end{equation}
where $\mathcal{D}(\cdot)$ is the cosine similarity function, $N$ is number of frames in the augmented sequence $\mathcal{S''}$. While~\cite{kwon2022context} uses distance instead of cosine similarity, in our experiments, cosine similarity resulted in a faster and more stabilized convergence during training.

We then predict the frame index, $\hat{j}$, in the original sequence $\mathcal{S}$, by weighing the frame indices with their
corresponding probabilities, as follows:

\begin{equation}\label{eq:predict}
  \hat{j} = \sum_{k=1}^{N} (\gamma_{j,k}\cdot g''(k)).
\end{equation}

The loss $\mathcal{L}_{\mathcal{S'} \mapsto \mathcal{S''}}$ can be computed as the mean squared error between the predicted frame index $\hat{j}$ and its ground truth frame index, which is preserved after augmentation:

\begin{equation}
   \mathcal{L}_{\mathcal{S'} \mapsto \mathcal{S''}} = \frac{1}{M}\sum_{j=1}^M \left\lvert g'(j)-\hat{j}\right\lvert ,
\end{equation}

To have a symmetrized loss we can similarly compute the  $\mathcal{L}_{\mathcal{S''} \mapsto \mathcal{S'}}$ as:

\begin{equation}
  \mathcal{L}_{\mathcal{S''} \mapsto \mathcal{S'}} = \frac{1}{N}\sum_{k=1}^N \left\lvert g''(k)-\hat{k} \right\lvert ,
\end{equation}
where $\hat{k}$ is the predicted frame index of frame $k$ in the original sequence $\mathcal{S''}$, where $s_k''\in \mathcal{S''}$, and $M$ is the number of frames in the augmented sequence $\mathcal{S'}$.
The final matching loss then can be defined as follows:

\begin{equation}\label{eq:loss}
  \mathcal{L}_{m} = \frac{1}{2}(\mathcal{L}_{\mathcal{S'} \mapsto \mathcal{S''}} + \mathcal{L}_{\mathcal{S''} \mapsto \mathcal{S'}}).
\end{equation}

\paragraph{Clip-level Implicit Clustering. }

Our implicit clustering component is inspired by \cite{chen2021exploring}, which introduces representation learning using a Siamese network architecture with image clustering. One side of the Siamese network is extended by introducing a clustering predictor, and a stop-gradient operation is applied on the other side. 
Simsiam~\cite{chen2021exploring}, however, is designed to work on a single image only and does not perform on videos.
We, therefore, extend its concept to clip-level clustering instead of frame-level clustering, by composing attention layers and a matching method to the clustering predictor for better frame alignment and action classification, unlike the original Simsiam~\cite{chen2021exploring}.
This helps to capture temporal context and leverage it in the domain for sequence alignment.  

The clip-level clustering predictor $H$ learns to map each output of the projection head $\mathcal{Z'}$ or $\mathcal{Z''}$ to a latent space, which represents its clustering prototype by maximizing the similarity between both latent vectors. 
At the same time, stop-gradient, which prevents gradient descent flow from the other latent vector of the clustering predictor, plays an essential role in preventing collapsing as the optimizer quickly finds a degenerated solution, and encourages the network to learn diverse and invariant features~\cite{chen2021exploring}. 
Our ablation study shows the impact of the stop gradient for the task of temporal sequence alignment. Similar to the matching loss, we also define a symmetrized loss as:

\begin{equation}
\begin{aligned}
  {\mathcal{L}_{c} = \frac{1}{2}(\mathcal{F}(\texttt{stopgrad}(\mathcal{Z'}),H(\mathcal{Z''}))} 
  +\\ \mathcal{F}(\texttt{stopgrad}(\mathcal{Z''}),H(\mathcal{Z'}))),
\end{aligned}
\label{eq:loss}
\end{equation}
where $\texttt{stopgrad}(\cdot)$ is the stop-gradient operation that prevents the gradient from being propagated from $\mathcal{Z'}$ and  $\mathcal{Z''}$ in the first and second terms, respectively.

After passing through $H$, the embeddings keep the temporal dimensionality and match between the clustering predictor and input sequence by matching function $\mathcal{F}(\cdot)$, which preserves temporal information, enabling H to predict clip-level prototype activities effectively. $\mathcal{F}(\cdot)$ calculates the negative cosine similarity based on matching indices in the original sequence as

\begin{equation}
\begin{aligned}
\mathcal{F}(\cdot \mid g', g'')(V, W) = \sum_{j=0}^{M}\sum_{k=0}^{N}\delta_{g'(j), g''(k)}\mathcal{D}(V_j, W_k),
\end{aligned}
\end{equation}

where $\delta_{x, y}$ is Kronecker delta function, which is 1 if $x=y$ and 0 otherwise.

Taking into account the disparity in the range of  $\mathcal{L}_{M}$ and $\mathcal{L}_{c}$, we optimize the following loss over the parameters of the networks as

\begin{equation}\label{eq:finalloss}
  \mathcal{L} = \mathcal{L}_{m} \cdot {\frac{2}{(1+ \epsilon)+\mathcal{L}_{c} }}. 
\end{equation}
$\mathcal{L}_{c}$ varies between $-1$ and $+1$, we therefore add $\epsilon$ (in practice we use $\epsilon = 1 \times 10 ^{-7}$)  to the denominator to ensure preventing division by zero. 
While $L_c$ prevents the latent space from collapsing across different activities, $L_m$ contributes to the sequence alignment, allowing the model to capture similarities between activities. 

At the inference, we feed the sequences without augmentation and utilize embeddings $\mathcal{U'}$ to evaluate our model.

\section{Evaluation}
In this section, we present and discuss the results of our evaluation on three datasets.
We first describe the datasets and the implementation details. Then, we describe the evaluation metrics used in this paper. We then show the performance of the proposed method on the tasks of Fine-grained Video Understanding and Action Recognition and compare it to the state-of-the-art methods. Finally, we demonstrate the results of an ablative analysis of our method.

\subsection{Datasets}
We evaluate our performance on three different datasets, PennAction~\cite{Zhang_2013_Penn}, IKEA ASM~\cite{Ben-shabat_2020_Ikea}, and H2O~\cite{Kwon_2021_H2O} following~\cite{kwon2022context,tran2023learning}.

\boldparagraph{H2O.}
The H2O dataset collects 3D hand poses and object poses while subjects perform a series of actions with a multi-view and an egocentric view. CASA~\cite{kwon2022context} selects the monotonic videos from one activity, \texttt{Pouring Milk}, and annotates the phases accordingly. The \texttt{Pouring Milk} activity contains 38 sequences, 10 phases. The longest sequence has a maximum of 865 frames. The entire dataset has 36 actions, 184 videos, and the longest sequence has a maximum of 1238 frames. 3D hand poses are provided in this dataset, therefore, similar to~\cite{kwon2022context} we use these poses as the source of input sequences.  

\boldparagraph{PennAction.}
The PennAction dataset is a collection of sports activities. For training and evaluation, we use the same 13 subset actions as~\cite {kwon2022context,tran2023learning,dwibedi2019temporal,haresh2021learning,liu2022learning}. These subset actions are divided into 2 to 6 phases, 371 videos, and up to 663 frames. To obtain the body poses for this dataset, we use FrankMocap~\cite{rong2021frankmocap}. 

\boldparagraph{IKEA ASM.}
The IKEA ASM dataset captures human activity during furniture assembly. Following~\cite{liu2022learning,kwon2022context,haresh2021learning,tran2023learning}, we train and evaluate on the \texttt{Kallax Drawer Shelf sequences}. They are divided into 17 phases and contain 90 videos up to 4078 frames. We obtain the body poses similar to the PennAction dataset. %

\subsection{Implementation Details}

\paragraph{Model Architecture.}
The model consists of the following network components: a multilayer perceptron (MLP), a self- and cross-attention module, a projection head, and a cluster predictor. The model aims to learn temporal context and alignment between augmented sequences.

\boldparagraph{MLP.}
We use a multilayer perceptron model (MLP) to transform the features from different modalities and align them to the same dimensionality. The MLP consists of three fully connected layers with ReLU activation. The input, hidden, and output layers are 1536, 512, and 128 for RGB modality, and have the same dimension as the input stream for 3D skeletons. For RGB modality, we use a pretrained network DINO~\cite{oquab2023dinov2} to extract features from the RGB images, and then pass them to the MLP to obtain the features in the same dimension as the corresponding dataset. For 3D skeleton modality, we use the hand keypoints for the H2O dataset $s_{\text{H2O}} \in \mathbb{R}^{17 \times 2 \times 3}$ and body keypoints $s_{\text{PennAction}} \in \mathbb{R}^{25 \times 3}$ and $s_{\text{IKEA~ASM}} \in \mathbb{R}^{25 \times 3}$ for PennAction and IKEA ASM, respectively. The MLP allows us to extract and reorder the features to fit into the network architecture and enhance the performance. As we use MLP, we set the length of the input as the maximum frame length of the dataset and perform zero-padding where the length is longer than the input sequence.

\boldparagraph{Self- and Cross-Attention.}
We have self- /cross-attention encoders as our encoder of the network. We set the 4 attention layers alternately to maximize temporal context learning capability. The keys and values are target features, and the queries are source features in the cross-attention layer.

\boldparagraph{Projection Head.}
We set 4 fully connected layers for Projection head. Different from CASA~\cite{kwon2022context}, we also incorporate batch-normalization inspired by SimSiam~\cite{chen2021exploring}, which is not essential, but leads to a slight increase in the performance empirically. 

\boldparagraph{Cluster Predictor.}
The cluster predictor consists of 2 self-attention layers. These attention layers consider temporal context while the module optimizes clustering prototypes.

\boldparagraph{Optimization.}
We apply the Adam optimizer~\cite{kingma2014adam} to update the model parameters. We use different batch sizes for each dataset: 32 for PennAction, 16 for H2O, and 4 for IKEA ASM. We start with a learning rate of $3 \times 10^{-3}$ and decay it by a factor of 2 every 50 epochs until 150 epochs. For the CASA baselines and the skeletal augmentation, we use their public code and parameters as described in~\cite{kwon2022context}.

\subsection{Evaluation Metrics}
We use the following evaluation metrics for the tasks of fine-grained video understanding. We do not use any labels for the training stage and only use them for the evaluation stage to compute the performance by training a linear classifier for the phase classification, and a linear regressor for the phase progress, as in the literature~\cite{kwon2022context,tran2023learning,dwibedi2019temporal}. 

\boldparagraph{Phase Classification} is per-frame classification. We train the model without labels, learn a linear classifier with phase labels (e.g., \texttt{bat swung back, bat hits ball}), and evaluate performance in the same activity.

\boldparagraph{Frame Retrieval} computes the average precision (AP), which shows the success rate of retrieving the frames with the same phase label in K frames using K-nearest neighbors in the embedding space.

\boldparagraph{Phase Progress} measures embedding effectiveness in capturing action progress.

\boldparagraph{Kendall's Tau} is a statistical measure of the alignment quality between two videos, which does not require any phase labels, unlike the other metrics.

\boldparagraph{Action Recognition} is per-clip classification. It measures the accuracy of the actions performed.  We evaluate it across the entire H2O and PennAction datasets. 
Unlike phase classification, which is limited by a monotonic sequence, action recognition is not bound by this restriction and can assess classification performance across multiple activities.

\subsection{Fine-grained Video Understanding}

\begin{figure}[t]
\begin{adjustbox}{width=0.8\columnwidth,center}
\begin{minipage}[c]{0.5\linewidth}
\includegraphics[width=\linewidth]{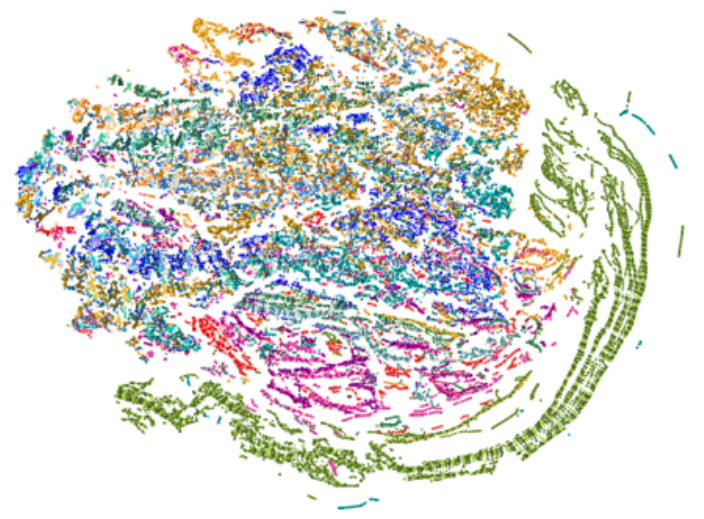}
\label{fig:casa}
\end{minipage}%
\begin{minipage}[c]{0.5\linewidth}
\includegraphics[width=\linewidth]{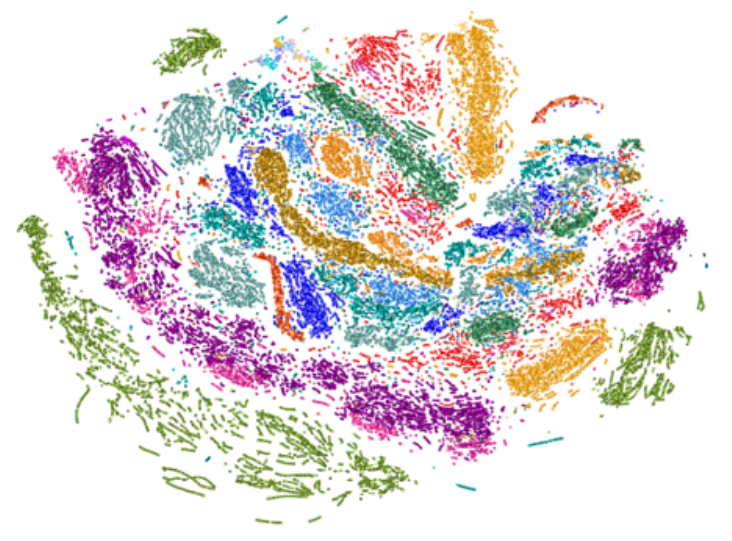}
\label{fig:masa}
\end{minipage}
\end{adjustbox}
  \vspace{-3mm}
\caption{\textbf{t-SNE visualization of the learned embedding trained on multi activity on PennAction.} (left) CASA shows embeddings fall into a single homogeneous space, while (right) ours can learn more generalizable and discriminative representations. Each color represents each activity.}
\label{fig:tsne}
  \vspace{-5mm}
\end{figure}

\begin{table*}
\caption{\small {\bf Performance comparison of phase classification, video progress, and Kendall's Tau}. The table illustrates superior performance across various datasets across most of the evaluation metrics for RGB modality. Additionally, for 3D skeletons, our proposed approach surpasses the state-of-the-art keypoints-based method.} 
\vspace{-5mm}
\begin{center}
\begin{adjustbox}{width=0.96\textwidth,center}
\begin{tabular}{lc|ccccc|ccccc|ccc}%
\toprule
& & \multicolumn{5}{|c}{PennAction~\cite{Zhang_2013_Penn}}&  \multicolumn{5}{|c}{H2O~\cite{Kwon_2021_H2O}}& \multicolumn{3}{|c}{IKEA ASM~\cite{Ben-shabat_2020_Ikea}} \\
 \multirow{2}{*}{Method} & \multirow{2}{*}{Modality}   &  \multicolumn{3}{c}{\% of Labels $\rightarrow$}   &  \multirow{2}{*}{Progress}& \multirow{2}{*}{$\tau$}  & \multicolumn{3}{c}{\% of Labels $\rightarrow$} &  \multirow{2}{*}{Progress}& \multirow{2}{*}{$\tau$}& \multicolumn{3}{c}{\% of Labels $\rightarrow$}  \\
\cline{3-5} \cline{8-10} \cline{13-15}
& &  10 & 50 &100& && 10 &50&100& && 10 &50&100\\ 

 \midrule

 LA2DS~\cite{tran2023learning}  & 2D heatmap & 89.27 & 92.30 & 92.63&0.9348&\underline{0.9887}&\underline{55.86}&64.58&70.12&\underline{0.9280}&0.9670&26.43&32.56&34.73\\
 LAV~\cite{haresh2021learning}  & 3D skeleton & 79.83 & 80.25 & 80.20 &0.6404&0.6983& 37.05&39.50&40.45& -& - &14.52&16.31&18.63\\
 TCC~\cite{dwibedi2019temporal} & 3D skeleton  & 79.53 & 83.75 & 84.51 &0.6268&0.6267&30.40&40.20&42.72&-&-&11.95&13.53&18.63\\
 CASA~\cite{kwon2022context}  & 3D skeleton & 88.55 & 91.87 & 92.20&\underline{0.9449}&0.9728&43.50&62.51&68.78&0.9086&0.9465&21.32&31.52&31.06\\

 MASA (ours) & 3D skeleton  &  89.26 & 92.46  & 93.03 & 0.9342 & 0.9585 & 45.94&\underline{69.97}&\underline{75.17}&\textbf{0.9400}&\underline{0.9674}&27.17&\underline{34.55}&\underline{36.87}\\

\hdashline
 TCC~\cite{dwibedi2019temporal} & RGB  & 79.72 & 81.11 & 81.35 &0.6638&0.7012&43.30&52.48&52.78&-&-&27.74&25.70&26.80\\
  TCN~\cite{sermanet2018time} & RGB & 81.99 &83.67&84.04 & 0.6762&0.7328	&-&-&-&-&-&25.17&25.70&26.80\\
 LAV~\cite{haresh2021learning} & RGB & 83.56 & 83.95 & 84.25 &0.6613&0.8047&35.38&51.66&53.43&0.5913&0.5323&\underline{29.78}&29.85&30.43\\

VAVA~\cite{liu2022learning}  & RGB & 83.89 & 84.23 & 84.48&-&-&-&-&-&-&-&\textbf{31.66}&33.79&32.91\\
 LA2DS~\cite{tran2023learning}  & RGB & 84.75 & 86.91 & 88.58&0.8891&0.9625&48.08&63.26&68.85&0.9113&0.9478&28.31&32.98&35.55\\
  VSP~\cite{zhang2023modeling}  & RGB & \underline{92.24} & \underline{92.48} & 93.12&0.9230&0.9860&-&-&-&-&-&-&-&36.09\\
 CARL~\cite{chen2022frame}  & RGB & - & - & \underline{93.18}&0.931	&0.9680&-&-&-&-&-&- &- &-\\
  
MASA (ours)  & RGB  & \textbf{92.40} &  \textbf{92.84}  & \textbf{93.63}  & \textbf{0.9914} & \textbf{0.9894} & \textbf{74.71} &\textbf{80.35} & \textbf{83.68} &0.9275&\textbf{0.9857}&24.31 & \textbf{35.18} &\textbf{38.83} \\

\hline
\end{tabular}
\end{adjustbox}
\end{center}

\vspace{-7mm}
\label{table:comparisons}
\end{table*}

We evaluate the performance of MASA in the context of fine-grained video understanding, comparing our results against state-of-the-art approaches. 
We report our results in two different modalities, RGB and 3D skeleton. Our method outperforms most of the metrics on both modalities.

Table~\ref{table:comparisons} presents a comprehensive overview of our comparative analysis. Notably, our method consistently achieves superior performance compared to previous approaches, particularly on the PennAction dataset and the H2O dataset. In the case of using 100\% labels for the linear classifier, our method outperforms state-of-the-art techniques by a substantial margin, exhibiting performance gains of 0.45\%, 2.74\%, and 13.56\% on PennAction,  IKEA ASM, and H2O, respectively. 
Notably, our method demonstrates significant performance gain for the H2O dataset for every portion of labels (10\%, 50\%, 100\%), which shows the effectiveness of our clip-level clustering for sequences that include multiple procedural steps to complete the pouring milk task.
Importantly, even when utilizing a reduced portion (10\% or 50\%) of labels for training, our method remains competitive or surpasses other methodologies. This underscores the robustness and generalizability of our approach across different datasets.

As shown in Table~\ref{table:comparisons}, the phase progress and Kendall's tau measure how well the sequences are aligned by phase and by mutual order. Our method achieves high scores on both metrics, thanks to the context module.
In Figure~\ref{fig:teaser}, we present qualitative results for sequence alignment. More qualitative results will be in the supplementary material.

\subsection{Generalization Capabilities}

We distinguish terminologies for clarity: 
\textbf{Action}: a single instance, e.g., \texttt{Baseball Swing}.
\textbf{Activity}: a series of actions that form a complete monotonic sequence.
\textbf{Phase}: a state defined by \textbf{key events} that mark significant transitions within the sequence. E.g., there are 3 phases ($P_{1,2,3}$) in \texttt{Baseball Swing}: $P_1$ $\rightarrow$ \texttt{Bat swung back} $\rightarrow$ $P_2$ $\rightarrow$ \texttt{Bat hit} $\rightarrow$ $P_3$. 
The existing methods for achieving optimal performance are mostly limited to aligning sequences of the same activity only and require separate models to be trained for each activity. In contrast, our approach demonstrates successful generalization to multiple activities.
To validate this generalizability, we conducted additional experiments using training on all activities on different tasks.

\paragraph{Multi Activity vs. Single Activity.}
We compare the performance of our approach when it is trained on all activities within the PennAction dataset, which has the highest number of sequences and activities among evaluation datasets, compared to training separate models for each individual activity.
To set up baselines, we select two state-of-the-art methods for each modality, CASA~\cite{kwon2022context} and CARL~\cite{chen2022frame}.
We retrained CASA with 3D skeletons on all activities and assessed its performance with the code that the authors provided. For the CARL baseline implementation, we also adopt the same code and parameters as specified in their paper to train a model per action.
As presented in Table~\ref{table:multi_act}, on 3D skeleton our proposed method exhibits a performance increase almost across all the metrics, which underscores the remarkable generalizability of our approach. 
In the case of Multi activity, CASA shows a significant drop in performance across all metrics compared to when trained on one activity. Whereas MASA on 3D skeleton only slightly decreases for 50\% and 100\% phase classification while keeping other metrics increased. 
The performance gain is more noticeable on RGB data. While CARL performs comparably or slightly better when using individual models per action, MASA stands out by outperforming all metrics when training a single model for all the activities simultaneously. This result demonstrates the effectiveness of clustering for sequence alignment, as it prevents embedding collapse and allows the model to leverage features shared across different activities. More comparisons are in the supplementary material.
Additionally, we visually compare the learned embeddings of CASA to ours when trained on multiple activities. As Figure~\ref{fig:tsne} illustrates the learned embeddings fall into a single homogeneous space across different activities, which results in difficulties in identifying different activities, while the learned embeddings by our method differentiate well between different activities.

\begin{table}[h]
\centering

\caption{\small \textbf{Comparison of our approach and CASA/CARL when trained on all activities of the PennAction dataset.} Despite a tenfold increase in the number of activities, our method demonstrates a significant performance improvement with the RGB modality and only a minor drop in performance with the 3D skeleton modality, highlighting its generalization to diverse activities.
}
\vspace{-3mm}
\begin{adjustbox}{width=1.0\columnwidth,center}
\begin{tabular}{@{}lcccccccc@{}}
\toprule
\multirow{2}{*}{Method} & \multirow{2}{*}{Mod.} & Multi- & \multirow{2}{*}{Progress} & \multirow{2}{*}{$\tau$} & \multirow{2}{*}{Retriev.} & \multicolumn{3}{c}{\% of Labels $\rightarrow$} \\
\cline{7-9}
&& Action & & & & 10 &50&100\\ 

 \midrule
CASA~\cite{kwon2022context}& 3D & $\times$ &  \textbf{0.9449}  & \textbf{0.9728} & \textbf{89.07}& \textbf{ 88.55} & \textbf{91.87}  & \textbf{92.20} \\
CASA~\cite{kwon2022context}& 3D & \checkmark &  0.9012  & 0.9236 & 84.95&  84.37 & 90.40  & 90.12 \\ \hdashline
MASA (ours)& 3D & $\times$ & 0.9341  & 0.9585 & 90.88& 89.26  & \textbf{92.46} & \textbf{93.03}  \\
MASA (ours)& 3D & \checkmark & \textbf{0.9906}  &\textbf{ 0.9652} &\textbf{ 90.96} & \textbf{90.16}  & 91.71  & 92.23 \\ \hdashline
CARL~\cite{chen2022frame}& RGB & $\times$ &  \textbf{0.931}  & 0.968 & 91.39&  - & -  & \textbf{93.18} \\
CARL~\cite{chen2022frame}& RGB & \checkmark & 0.918  & \textbf{0.985} & \textbf{91.82} &  - & -  & 93.07 \\  \hdashline
MASA (ours) & RGB & $\times$ &  0.9103 & 0.9646  & 91.56 & 90.43   & 91.25  & 92.12   \\
MASA (ours) & RGB  & \checkmark & \textbf{0.9914}   & \textbf{0.9894}& \textbf{93.07} & \textbf{92.40} & \textbf{92.84}   & \textbf{93.63} \\

\bottomrule
\end{tabular}
\end{adjustbox}

\label{table:multi_act} 
\vspace{-3mm}
\end{table}

\begin{table}[h]
\caption{\small  \textbf{Comparing our method's performance in action recognition with RGB images, 3D hand poses, and their combination.} It demonstrates the effectiveness of our method across different modalities.}
\vspace{-3mm}
\centering
\begin{adjustbox}{width=1.0\columnwidth,center}
\begin{tabular}{@{}lllcc@{}}
\toprule
Dataset & Method & Modality  & SSL & Acc. (\%) \\
\midrule
\multirow{8}{*}{H2O~\cite{Kwon_2021_H2O}}
&Linear classifier & 3D skeleton & -  & 33.88 \\
&CASA & 3D skeleton & \checkmark  & 47.52 \\
&TA-GCN~\cite{Kwon_2021_H2O}& 3D skeleton & $\times$ & 58.92\\
&Ours & 3D skeleton & \checkmark &  57.03  \\

&Wen et al.~\cite{wen2023hierarchical}& RGB & $\times$ & \textbf{86.36}\\

&Linear classifier & RGB & -  & 79.34 \\ 
&MASA (ours)& RGB  & \checkmark & 81.81\\ %
&MASA (ours)& 3D skeleton \& RGB & \checkmark & 82.64 \\ 

\midrule
\multirow{6}{*}{PennAction~\cite{Zhang_2013_Penn}} 
&Linear classifier & 3D skeleton & -  & 82.36 \\
&CASA & 3D skeleton & \checkmark  & 90.96 \\
&MASA (ours) & 3D skeleton & \checkmark &95.80 \\
&Linear classifier & RGB & -  & 91.50 \\
&MASA (ours) & RGB &  \checkmark &96.73 \\
&MASA (ours) & 3D skeleton \& RGB & \checkmark & \textbf{97.03} \\
\bottomrule
\end{tabular}
\end{adjustbox}

\label{tbl:multimodality}
\vspace{-5mm}

\end{table}

\paragraph{Action Recognition with Different Modalities.}

We evaluate the effectiveness of MASA for the task of action recognition. Action recognition also allows us to measure the performance of clustering across different activities. The purpose of this evaluation is to observe the generalization ability across various actions.
For this task, we use action recognition labels from the H2O dataset and repurpose phase labels (atomic action) as action labels in the PennAction dataset to create more diverse action classifications. For the evaluation, we uniformly sample 16 frames per action and run a linear classifier with the training labels to maintain a self-supervised manner.
As a baseline, we run a linear classifier using 3D hand poses and RGB images instead of learned embeddings. 
MASA shows better action recognition performance across two datasets and modalities compared to the linear classifier and CASA baselines.

To compare between modalities, as RGB modality compared to 3D hand skeleton modality provides a richer representation, we achieve a significant improvement of 24.78\% and 0.93\%  in accuracy on H2O and PennAction, respectively.
Furthermore, we also address multi modality for action recognition. 
We fuse 3D hands and RGB images together by concatenating their features before inputting them into our context-aware module. The results presented in Table~\ref{tbl:multimodality} highlight the effectiveness of our approach across different modalities.
 When combining 3D hand skeletons and RGB data, our results demonstrate great performance levels comparable to state-of-the-art supervised learning-based methods~\cite{wen2023hierarchical,cho2023transformer} even though we only use a simple linear classifier to recognize actions. This shows the potential of our method in addressing general action recognition tasks.

\subsection{Ablation Studies}

\begin{table}[h]
\caption{\small \textbf{Impact of component exclusions on evaluation metrics on the H2O dataset.} Each component contributes to the performance, proving our design choice. }
\vspace{-3mm}
\centering
\begin{adjustbox}{width=1.0\columnwidth,center}
\begin{tabular}{@{}l|ccccccc@{}}
\toprule

  \multirow{2}{*}{w/o} &Self & Stop &  Dual & Trim &Cluster &Cross  & \multirow{2}{*}{-}\\ 
 & att.& gradient & aug. & aug. &predictor  & att.   &  \\ 
\midrule
  Class. & 74.98&76.68 &77.68& 78.87&79.73 &79.33 &83.68\\
  
  Progress&0.9335& 0.9173 &0.9092 & 0.9192&0.9202 &0.9226&0.9275\\
  Tau &0.9630& 0.9712&0.9803 & 0.9824 &0.9833&0.9751  &0.9857\\

\bottomrule
\end{tabular}
\end{adjustbox}
\label{table:ablation_network}
\vspace{-3mm}
\end{table}

We ablate MASA on the H2O (\texttt{Pouring Milk}) dataset. The results on phase classification, phase, progress, and Kendall's Tau metrics are shown in Table~\ref{table:ablation_network}. 
Our study confirms the importance of the stop gradient, the cluster predictor, and trimming for the clustering task.  
Notably, the exclusion of stop gradient leads to a significant drop of 7.0\% in classification accuracy. The collapse of embeddings happens with the clustering predictor in this case.
Moreover, omitting either of the attention modules results in a decrease in accuracy by about 4.35\% and 8.7\% for cross-attention and self-attention in classification, respectively. This indicates that the attention module plays an important role in learning the representations in the context-aware module.

\section{Conclusion}
In this paper, we present MASA, a novel framework for self-supervised sequence alignment that can handle a wide range of activities and scenarios. The proposed framework introduces clip-level implicit clustering and identification of prototypes for different activities in the latent space as an additional constraint in the optimization schema for the task of temporal video alignment. We also propose a dual augmentation technique, which can learn more generalizable and discriminative representations than existing methods. We demonstrated the superiority of MASA on three diverse datasets compared to the state-of-the-art methods and provided ablation studies and quantitative analysis. Naturally, the precision of this method may be impacted, especially for the very long videos as we showed in our experiments, which can be considered as a future work direction.
Nevertheless, our approach is not limited to RGB only and can leverage other modalities such as skeletons as we demonstrated.
Overall, this paper makes significant contributions to the field of temporal sequence alignment and has the potential to impact a wide range of applications.

{
    \small
    \bibliographystyle{ieeenat_fullname}
    \bibliography{main}
}

\end{document}


\maketitle

\setcounter{page}{1}

\newcommand{\beginsupplement}{%
        \setcounter{table}{0}
        \renewcommand{\thetable}{S\arabic{table}}%
        \setcounter{figure}{0}
        \renewcommand{\thefigure}{S\arabic{figure}}%
        \setcounter{section}{0}
        \renewcommand{\thesection}{S}%
     }
\beginsupplement
In this supplementary material, we first elaborate on our implementation details, such as the parameter, augmentation, image backbone, and pose estimator settings. 
We then demonstrate the results on the task of frame retrieval to show the success rate of retrieving the frames with the same phase label.
Furthermore, we present qualitative results using confusion matrices to have a better understanding of action recognition capabilities through our method.
Finally, we show more qualitative results, including frame retrieval and sequence alignment on both RGB and 3D skeleton modalities.

\subsection{Implementation Details}
\boldparagraph{Parameter Settings.}

\begin{table}[h]
\begin{center}
\begin{adjustbox}{width=0.8\columnwidth}
\begin{tabular}{@{}llll@{}}
\toprule
Hyperparameter & \multicolumn{3}{l}{Value} \\
\midrule
Optimizer & \multicolumn{3}{l}{ADAM}\\
Temperature ($\lambda_{temp}$) &\multicolumn{3}{l}{0.1}\\
3D geometric noise probability & \multicolumn{3}{l}{30 \%} \\
Noise standard deviation ($\sigma$) & \multicolumn{3}{l}{$10^{\circ}$ (angle), 0.1 (VPoser, translation)}\\
Number of attention layers ($N_{att}$)& \multicolumn{3}{l}{4}\\
\midrule
Datasets & Penn & H2O & IKEA \\
\midrule
Batch Size & 64 & 32 & 16  \\
Input dimension &  75 & 51 & 75\\
Number of heads (parallel attention layers)&  15 & 17 & 15 \\
Learning rate & $3 \times 10^{-3}$ & $1.5 \times 10^{-3}$ &  $7.5 \times 10^{-4}$  \\
Learning rate (all) &  $3 \times 10^{-4}$ & $1.5 \times 10^{-4}$ & - \\

\bottomrule
\end{tabular}
\end{adjustbox}
\end{center}
\caption{\small {\bf Hyperparameters in our experiment.}}
\label{table:hyperparameters}
\end{table}

In Table~\ref{table:hyperparameters}, we list hyperparameters that we use %
for training our models. We will also release our code upon acceptance of the paper.

\boldparagraph{Augmentations.} 
In accordance with the methodology proposed by CASA~\cite{kwon2022context}, we employ spatial augmentations to enhance the robustness of our model. Specifically, we utilize four spatial augmentations (angle, flip, translation, and VPoser) for the PennAction and IKEA~ASM datasets and three spatial augmentations (angle, flip, and translation) for the H2O dataset. The noise parameters associated with these augmentations are detailed in Table~\ref{table:hyperparameters}. Notably, our method diverges from CASA in terms of temporal augmentations. 
Unlike CASA, our approach contains a temporal cropping method that helps to learn partially trimmed sequence clustering and our proposed Dual Augmentation technique, which can cover slower augmentation.

\boldparagraph{Image Backbone Settings.}
For this image backbone, we use the pretrained DINO~\cite{oquab2023dinov2} model. DINO is a vision transformer-based model, trained using a self-supervised learning method on 142 million unlabeled images. We used a distilled version of the original model, which generates visual features with a dimensionality of 1536.

\begin{figure}
\centering
   \fbox{\includegraphics[width=0.7\columnwidth]{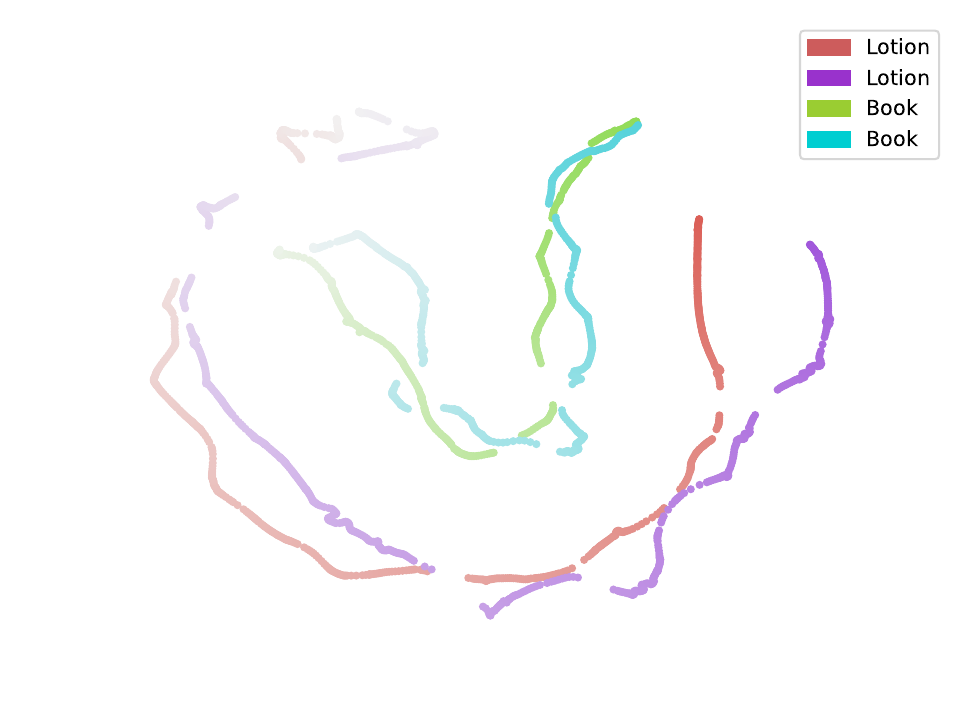}}
   \captionof{figure}{{\textbf{The t-SNE visualization of different activities in the H2O dataset.} \textcolor{blue}{Blue} and \textcolor{green}{green} are book activity sequences while \textcolor{red}{red} and \textcolor{violet}{violet} are 
   lotions sequences. Intra-class sequences of books and lotions exhibit strong alignment, whereas inter-class separation is pronounced in the embedding space. Less opacity means early in the sequence. } 
   } 
   \label{fig:separataion}
\end{figure}

\boldparagraph{Pose Estimator Settings.} 
We use 3D human body pose from SMPL~\cite{Pavlakos2019SMPL-X} and hand pose from MANO~\cite{Romero2017MANO} by following CASA~\cite{kwon2022context}. Both SMPL and MANO models take pose parameters $\theta_{smpl} \in \mathbb{R}^{72}$, $\theta_{mano} \in \mathbb{R}^{51\times2}$ and shape parameters $\beta_{smpl} \in \mathbb{R}^{10}$, $\beta_{mano} \in \mathbb{R}^{10\times2}$ and return 3D keypoints for body $s_{penn} \in \mathbb{R}^{25\times3}$ and hand $s_{hand} \in \mathbb{R}^{42\times3}$. %

\begin{table}[t]
\begin{center}
\caption{\small {\bf Comparison of frame retrieval performance.} 
Our framework outperforms other methods on the H2O and PennAction datasets for frame retrieval.}
\label{table:finegrained_retrieval}
\begin{adjustbox}{width=0.8\columnwidth,center}
\begin{tabular}{@{}lllccc@{}}
\toprule
Datasets & Method & Modality & AP@5 & AP@10 & AP@15 \\
\midrule
\multirow{11}{*}{PennAction~\cite{Zhang_2013_Penn}} 
&SAL~\cite{misra2016shuffle}& RGB& 76.04&75.77&75.61  \\
&TCN~\cite{sermanet2018time}& RGB&77.84&77.51&77.28  \\
&TCC~\cite{dwibedi2019temporal}&RGB & 76.74&76.27&75.88  \\
&LAV~\cite{haresh2021learning}&RGB &79.13&78.98&78.90  \\
& VAVA~\cite{liu2022learning}  & RGB & 81.52 & 80.47 & 80.67\\
&LA2DS~\cite{tran2023learning}   & 2D heatmap &\textbf{93.07} & 91.84& 91.35\\
&CASA~\cite{kwon2022context}& 3D skeleton&89.90 & 89.44 & 89.07 \\
&CARL~\cite{chen2022frame} & RGB & 92.28& \underline{92.10} & \underline{91.82}\\
&VSP~\cite{zhang2023modeling} & RGB & 92.56 & -& -\\
&MASA (ours) & 3D skeleton &  91.19 & 91.03 &  90.88  \\
&MASA (ours) & RGB & \underline{93.01} & \textbf{92.91}  & \textbf{92.82}\\
\hline
\multirow{10}{*}{IKEA ASM~\cite{Ben-shabat_2020_Ikea}}
& SaL~\cite{misra2016shuffle} & RGB & 15.15 &14.90&14.72	\\
& TCN~\cite{sermanet2018time} & RGB & 19.15 &19.19&19.33	\\
& TCC~\cite{dwibedi2019temporal} & RGB& 19.80&  19.64&  19.68  \\
&LAV~\cite{haresh2021learning} &RGB  & 23.89 & 23.65 & 23.56\\
& VAVA~\cite{liu2022learning}  & RGB & \underline{29.58} & 28.74 & 28.48\\
& LA2DS~\cite{tran2023learning}   & 2D heatmap& \textbf{32.44} & \textbf{31.89} & \textbf{31.56}\\
& CASA~\cite{kwon2022context}    & 3D skeleton & 28.92 & \underline{28.88} & \underline{28.61} \\
&VSP~\cite{zhang2023modeling} & RGB & 26.54 &-  &-\\
& MASA (ours)  & 3D skeleton  &  25.44  & 24.57 & 24.55 \\ 
& MASA (ours)  & RGB  & 23.77  & 24.07  & 24.28 \\ \hline
\multirow{5}{*}{H2O~\cite{Kwon_2021_H2O}}
&LAV~\cite{haresh2021learning}   & RGB&47.55& 45.56&44.61\\
& LA2DS~\cite{tran2023learning}   & 2D heatmap& 67.51 & 63.11 & 61.75\\
&CASA~\cite{kwon2022context} & 3D skeleton &60.13 & 59.44&59.01 \\
& MASA (ours)  & 3D skeleton  & \underline{66.79} &  \underline{66.18} & \underline{66.30}\\ 
& MASA (ours)  & RGB & \textbf{78.92}  & \textbf{78.55}  & \textbf{78.22}  \\ 

\hline

\hline
\end{tabular}

\end{adjustbox}
\end{center}

\end{table}

\subsection{Action Recognition Qualitative Results}
\begin{figure*}
   \includegraphics[width=0.87\textwidth]{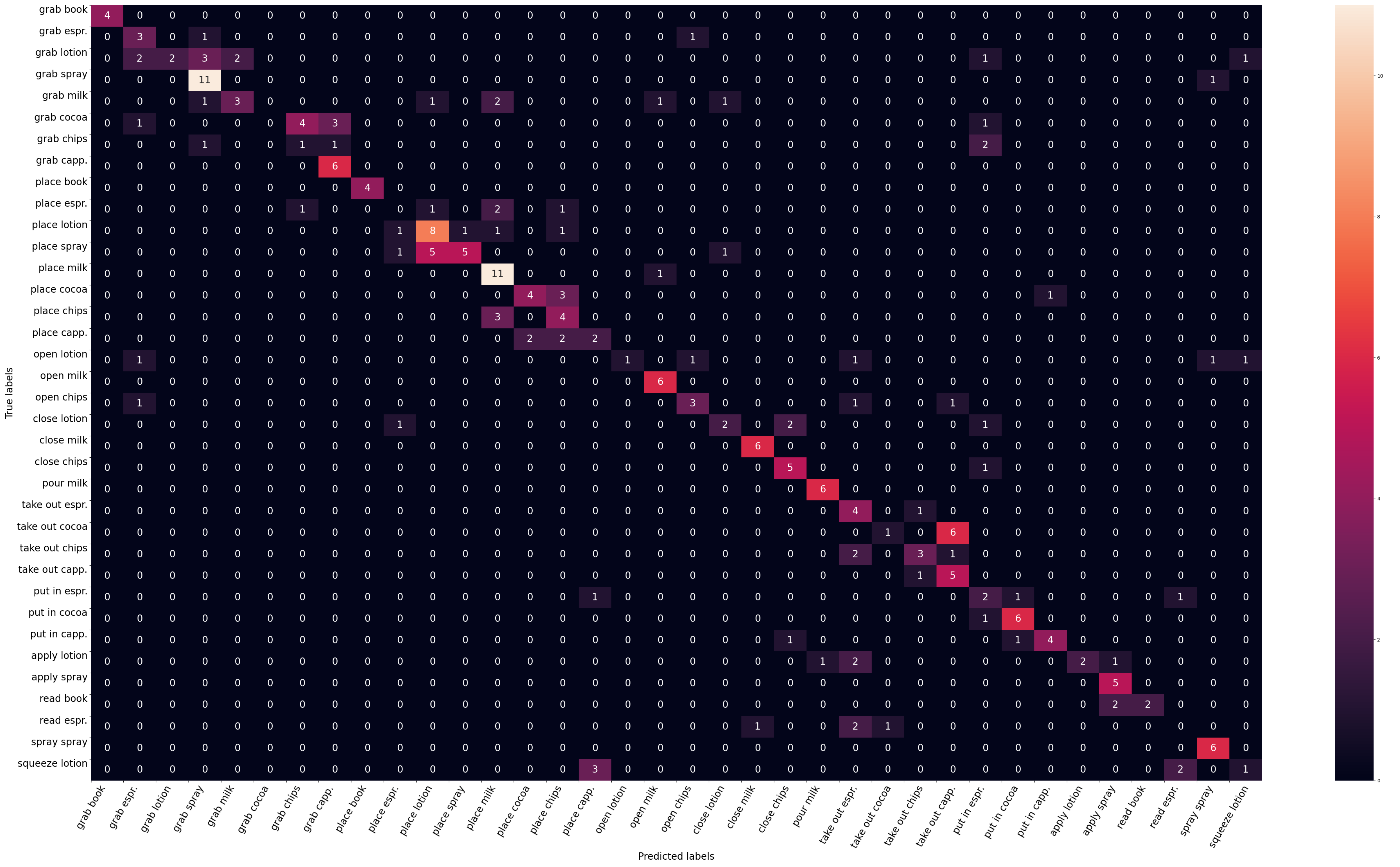}
   \captionof{figure}{{\textbf{Action Recognition only with hand poses on the H2O dataset.} The confusion matrix shows MASA successfully classifies action labels. However, the lack of texture with hand pose only leads to suboptimal performance in classifying nouns.} 
   } 
   \label{fig:action_pose}
\end{figure*}

\begin{figure*}
\includegraphics[width=0.87\textwidth]{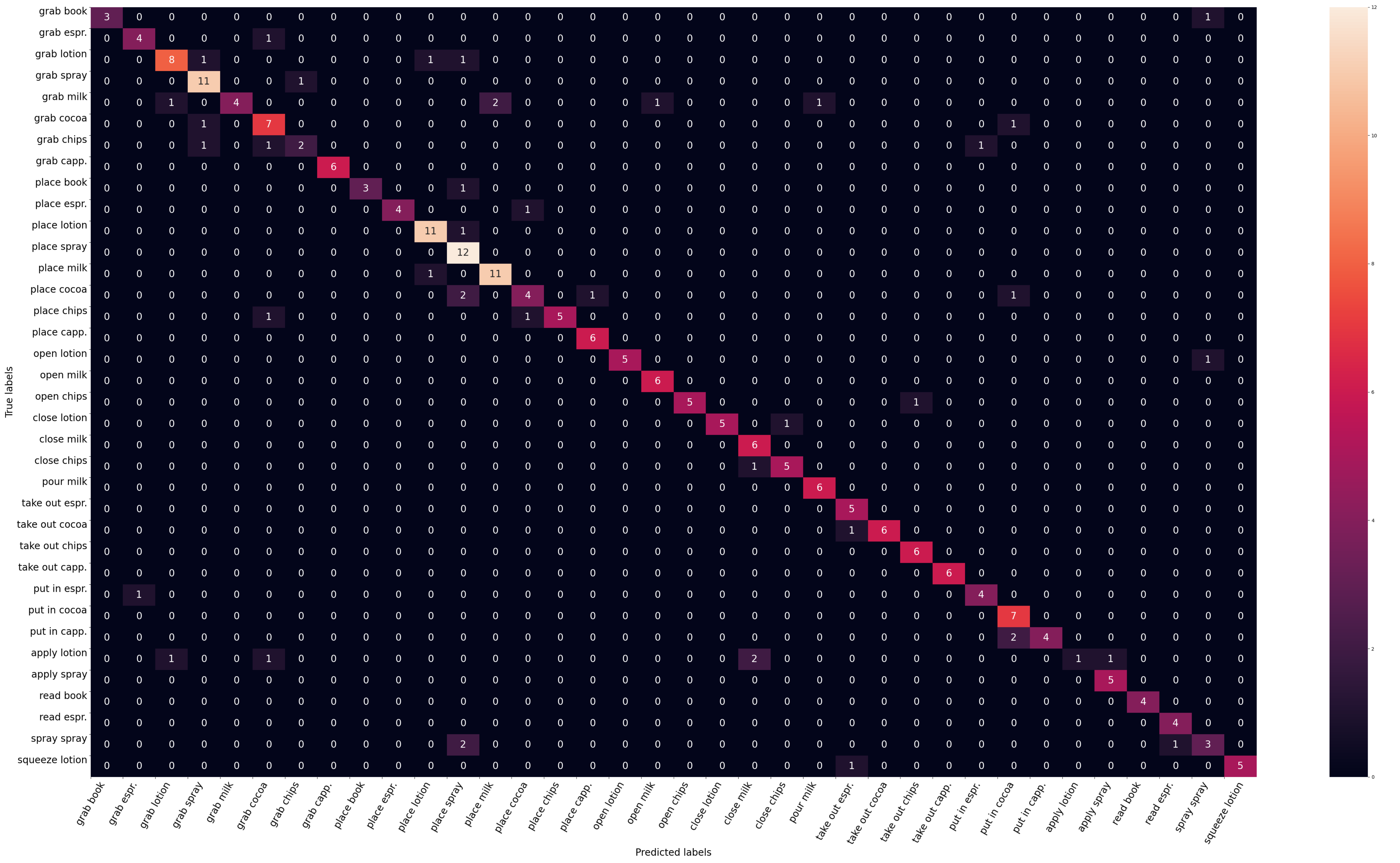}
   \captionof{figure}{{\textbf{Action Recognition with hand poses and RGB images on the H2O dataset.} Together with hand poses and RGB images, MASA achieves high performance across all actions. It shows the adaptability of our proposed framework to multimodal data.} 
   } 
   \label{fig:action_pose_dino}
\end{figure*}

We illustrate the action recognition confusion matrices pertaining to the H2O dataset in Figure~\ref{fig:action_pose} and Figure~\ref{fig:action_pose_dino}. 
Figure~\ref{fig:action_pose} demonstrates the confusion matrix generated solely from hand pose inputs on the H2O dataset. Our method, based on 3D skeletal information, engages in action recognition by exploiting hand poses. However, it is noteworthy that relying exclusively on hand pose information presents a challenge in distinguishing between nouns, such as  \texttt{Lotion} and \texttt{Spray}.

To address this challenge, we incorporate both hand poses and RGB images, as depicted in Table 4 of the main paper. Figure~\ref{fig:action_pose_dino} shows the network's enhanced ability to classify actions with the inclusion of additional texture information. This outcome indicates the efficacy of the frame alignment pretext task through multi-modal self-supervised learning for the task of action recognition.

\subsection{Frame Retrieval}
\begin{figure*}
   \includegraphics[width=\textwidth]{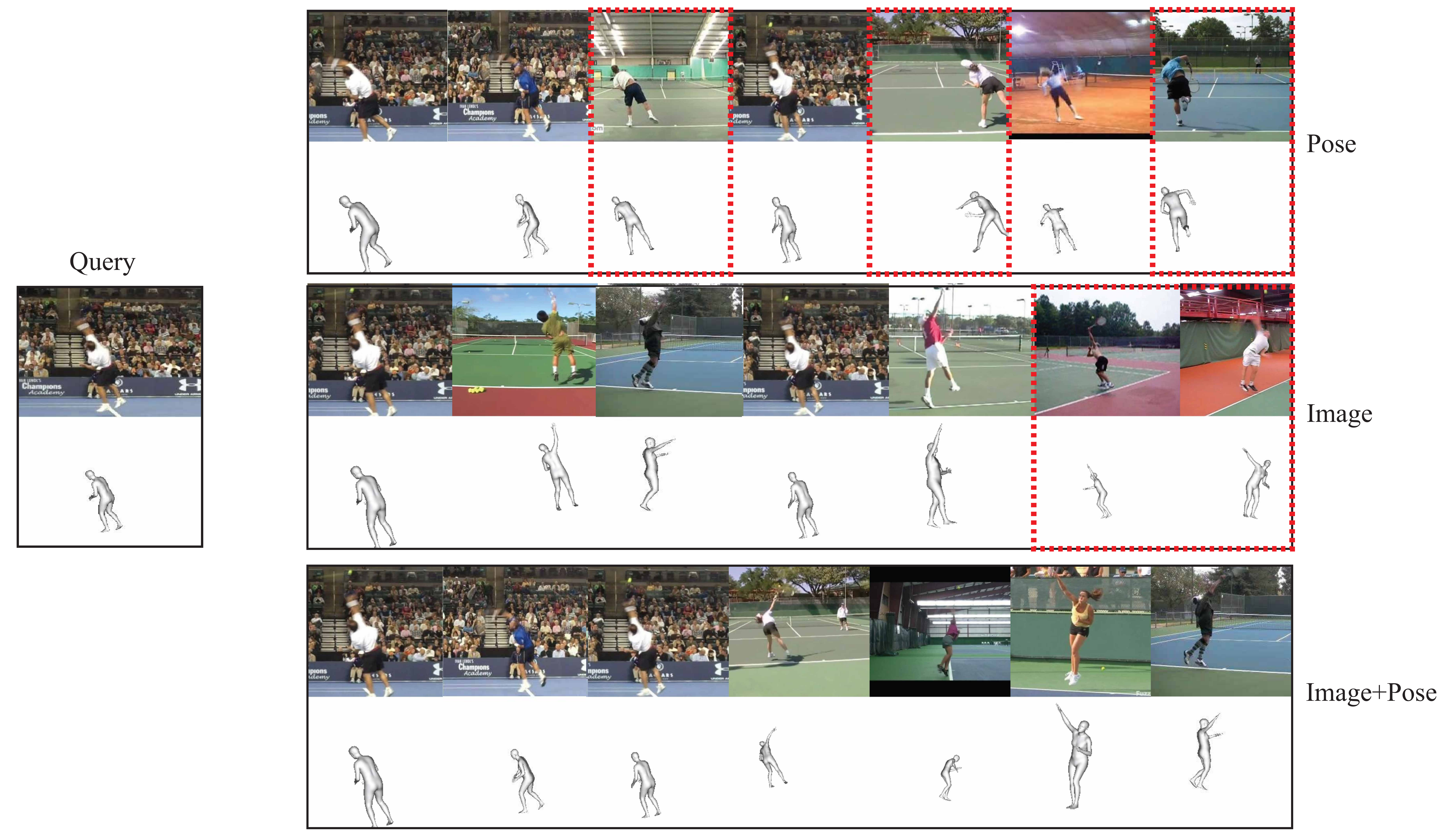}
   \captionof{figure}{{\textbf{Frame retrieval results.} We present the frame retrieval results using different modalities: pose only, image only and both pose and image together (from top to bottom). The frames that are incorrectly retrieved are marked with red dashed lines. MASA retrieves more successfully when it fuses different modalities, as they complement each other. These frame retrieval results demonstrate the ability of MASA to generalize to different modalities and show its potential for multi-modal sequence alignment.
   } 
   } 
   \label{fig:retrieval}
\end{figure*}

We present the frame retrieval performance in Table~\ref{table:finegrained_retrieval}. Our method demonstrates strong performance on the PennAction and H2O datasets. 
On the IKEA ASM dataset, however, our fine-grained retrieval results show a lower level of efficacy due to the complexity of optimization in the long and non-monotonic sequences for frame retrieval.
 However, as shown in Table 1, MASA outperforms others in phase classification, highlighting the strong potential of our representations.
Also, LA2DS has a higher dimensionality in the skeleton modality than CASA and MASA. We present the frame retrieval qualitative results in Figure~\ref{fig:retrieval}.

\begin{table}
\caption{\small More results for multi activity training on PennAction. Most of the methods show significant performance drops  when training multiple activities on one model (see Tab. 1).}
\begin{center}
\begin{adjustbox}{width=1.0\columnwidth,center}
\begin{tabular}{@{}llcccccccc}
\toprule
 &Method &  SAL~\cite{misra2016shuffle} & GTCC~\cite{donahue2024learning} & TCC~\cite{dwibedi2019temporal} & TCN~\cite{sermanet2018time} & LAV~\cite{haresh2021learning} & VAVA~\cite{liu2022learning}  & CARL~\cite{chen2022frame} & MASA (ours) \\
\midrule

\multirow{3}{*}{Single act.} 
&Phase Class. & 79.96& 81.30  & 81.35 & 84.04&  84.25& 84.48 & 93.18 & 92.12 \\
&Progress& 0.5943& 0.7080 & 0.6638 & 0.6762 & 0.6613 & 0.7090  & 0.9310 & 0.9103 \\
&$\tau$& 0.6336 & 0.7012 & 0.8830 & 0.7328 & 0.8047 & 0.8050 & 0.9680 & 0.9646 \\
\hdashline

\multirow{3}{*}{Multi act.}
&Phase Class. & 68.15 & 86.70 &74.39 & 68.09 &  78.68& 80.30 & \underline{93.07} & \textbf{93.63} \\
&Progress& 0.3903 & 0.8550 & 0.5914 & 0.3834 & 0.6252 & 0.6480 & \underline{0.9180} & \textbf{0.9894} \\
&$\tau$& 0.4744 & 0.9490 & 0.6408 & 0.5417 & 0.6835 & 0.7620 & \underline{0.9850} & \textbf{0.9914} \\
\hline

\end{tabular}
\end{adjustbox}
\end{center}
\label{table:multi_results}
\end{table}

\subsection{Multi Activity Comparisons}
Table \ref{table:multi_results} together with extensive experiments presented in our main paper supports the effectiveness of our method. The methods other than GTCC drop the phase classification performance significantly (7.43\%) when trained on multiple activities. GTCC and MASA are exceptions, showing improved performance under mult activity settings. However, GTCC does not achieve optimal performance relative to MASA (by 6.93\%) compared to the other baseline methods. Please note that most of the methods in the table also require action labels to train unlike MASA, which learns the representation purely without any labels.

\subsection{Potential Societal Impact}
While our research contributes significantly to the advancement of AR, VR, and video understanding applications, it has the potential for misuse in the surveillance and monitoring of people. Such misuse raises valid privacy concerns and should be guided by ethical AI principles.

\subsection{Responsibility to Human Subjects Data}  
Based on our current understanding, the IKEA ASM and H2O datasets are collected in a controlled lab environment, involving carefully selected subjects. Additionally, we received confirmation from the authors of the H2O dataset that they followed proper procedures during data collection, including obtaining agreements from subjects and securing IRB approval. The PennAction dataset comprises publicly available online resources such as content from YouTube. Therefore, to the best of our knowledge, using the datasets does not contravene privacy regulations.

\subsection{Qualitative Results}
\begin{figure*}
   \includegraphics[width=\textwidth]{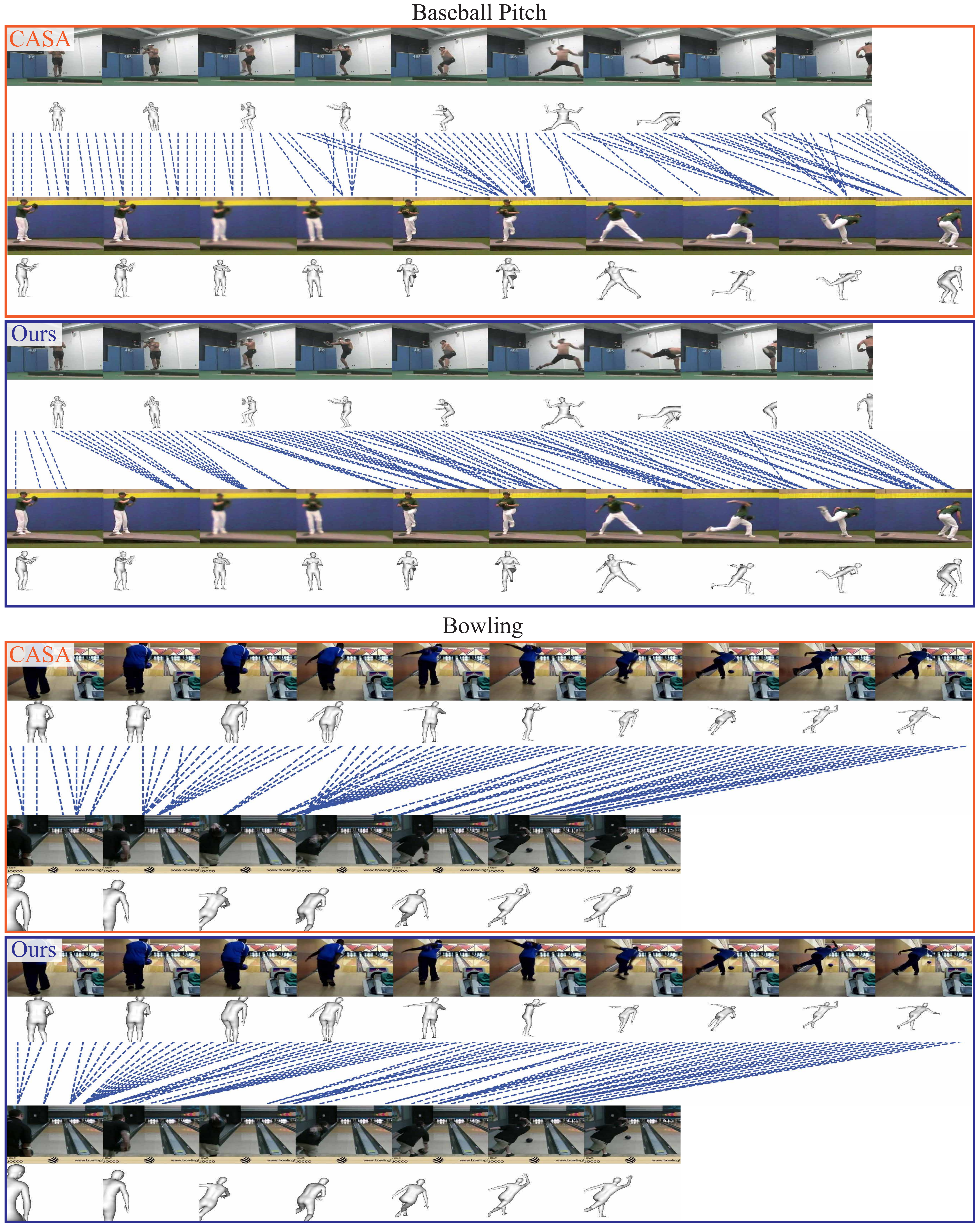}
   \captionof{figure}{{\textbf{Sequence alignment results.} MASA shows better alignment compared to the CASA method. For instance, in the baseball pitch example, we can observe more invalid matches that violate temporal consistency, whereas our method is more temporally consistent.} 
   } 
   \label{fig:alignment1}
\end{figure*}

\begin{figure*}
   \includegraphics[width=\textwidth]{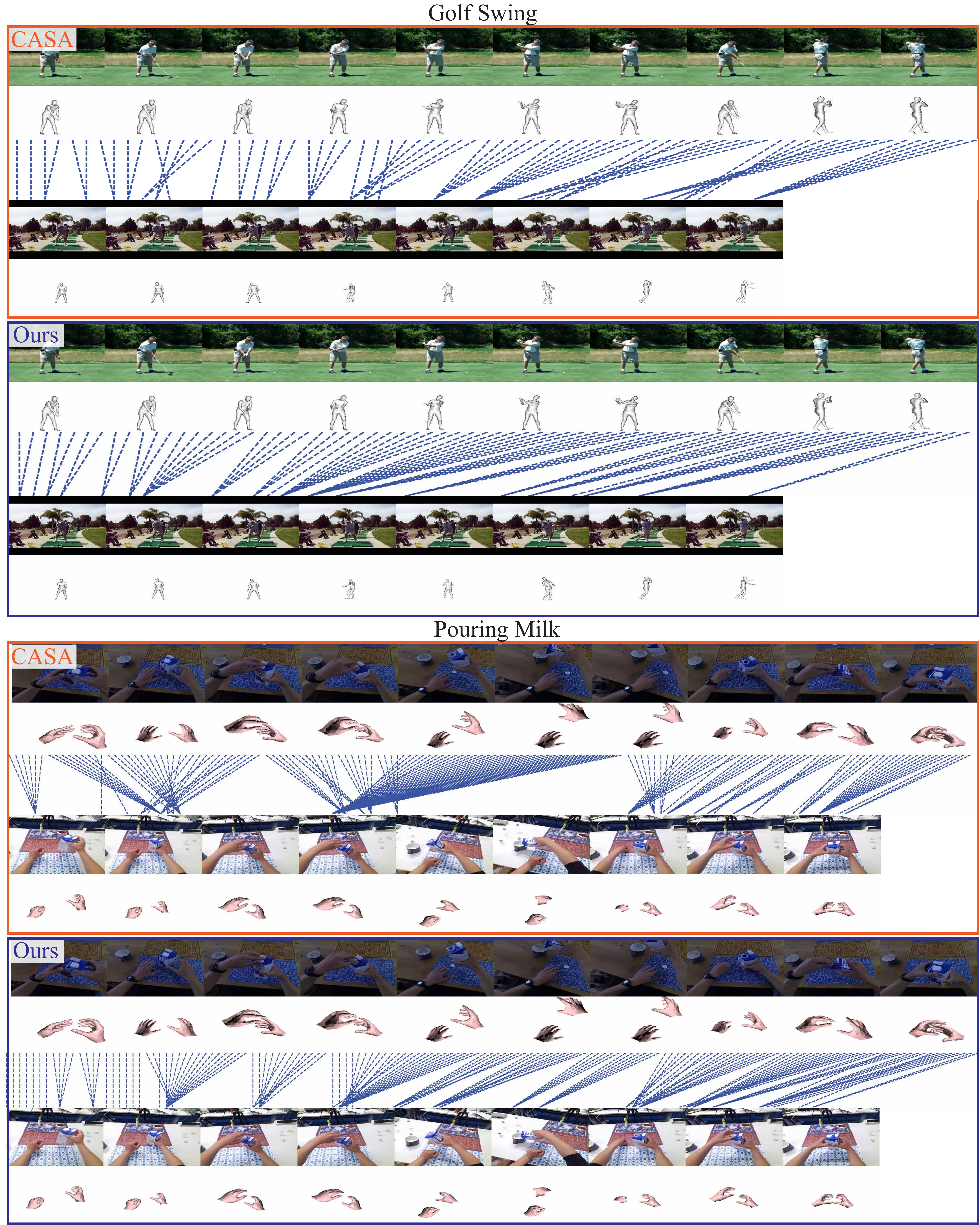}
   \captionof{figure}{{\textbf{Sequence alignment results.} Even in longer sequences on the H2O dataset compared to the PennAction dataset, MASA achieves robust results in alignment.} 
   } 
   \label{fig:alignment2}
\end{figure*}

\begin{figure*}
   \includegraphics[width=\textwidth]{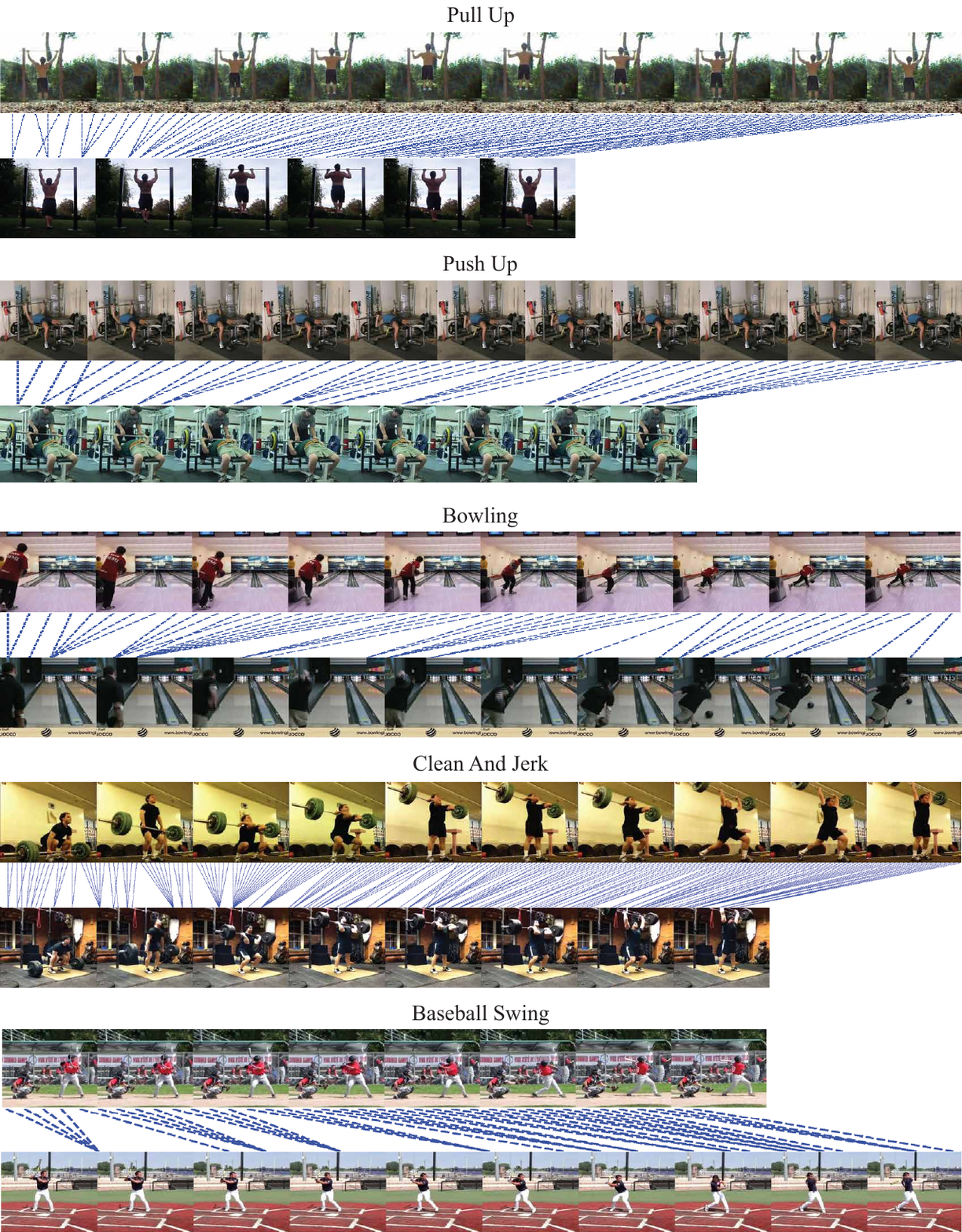}
   \captionof{figure}{{\textbf{Sequence alignment results.} MASA with RGB modality shows strong and robust results for the sequence alignment task.} 
   } 
   \label{fig:alignment3}
\end{figure*}

Figure~\ref{fig:alignment1} and \ref{fig:alignment2} show framewise sequence alignment qualitative results compared to CASA~\cite{kwon2022context} using 3D skeleton modality. To demonstrate our generalization ability with multiple activities, we use models trained on all the activities in the PennAction and H2O datasets, respectively, for visualization. 
MASA is able to align frames in different sequences more accurately with the multi activity training setting. 
Figure~\ref{fig:alignment3} shows alignment results using RGB modality. The results demonstrate a strong alignment performance of MASA with the high dimensional data such as RGB images.

\clearpage
\clearpage

{
    \small
    \bibliographystyle{ieeenat_fullname}
    \bibliography{main}
}